# Multifidelity Surrogate Models: A New Data Fusion Perspective

Daniel N. Wilke[1]

[1] Emerging Engineering Technology (EET) Lab, Department of Mechanical Engineering, University of Pretoria, Pretoria, Hatfield, 0086
Email: nico.wilke@up.ac.za

**Abstract:**

Multifidelity surrogate modelling combines data of varying accuracy and cost from different sources. It strategically uses low-fidelity models for rapid evaluations, saving computational resources, and high-fidelity models for detailed refinement. It improves decision-making by addressing uncertainties and surpassing the limits of single-fidelity models, which either oversimplify or are computationally intensive. Blending high-fidelity data for detailed responses with frequent low-fidelity data for quick approximations facilitates design optimisation in various domains.

Despite progress in interpolation, regression, enhanced sampling, error estimation, variable fidelity, and data fusion techniques, challenges persist in selecting fidelity levels and developing efficient data fusion methods. This study proposes a new fusion approach to construct multi-fidelity surrogate models by constructing gradient-only surrogates that use only gradients to construct regression surfaces. Results are demonstrated on foundational example problems that isolate and illustrate the fusion approach's efficacy, avoiding the need for complex examples that obfuscate the main concept.

## 1. Introduction

Multifidelity surrogate modelling aims to combine data from various sources that may have varying accuracy and computational costs. The key idea is to use lower-fidelity models for many quick evaluations, conserving computational resources while incorporating high-fidelity models to refine predictions. This approach also helps to address uncertainties to enable more informed decision-making. It can overcome the limitations of single-fidelity models, which may oversimplify real-world physics or be too expensive to compute tractably, with Toal (2015) presenting some guidelines. Multifidelity surrogate modelling expedites design optimisation in engineering fields, including aerospace and automotive design, and trades precision for computational effort in materials science. Financial multifidelity models help manage risk more effectively by combining data from various sources of reliability. Moreover, it supports environmental modelling, enabling more accurate predictions for climate change and natural disaster simulations.

Zhang et al. (2018) and Fernández-Godino et al. (2019) employed linear regression to correct multifidelity surrogate models, underscoring the method's effectiveness in handling complex engineering problems. These studies, along with contributions from researchers like Choi et al. (2005), Robinson et al. (2008), and Yamazaki et al. (2010), showcase the application of these models in design optimisation and predictive analysis. The research demonstrates how multi-fidelity surrogate modelling can enhance the efficiency of reduced-order models, as shown by Lee et al. (2016). It provides strategies for mixed-variable optimisation using models, as Singh and Grandhi (2010) explored. Zimmermann et al. (2010) and Han et al. (2013) have explored gradient-enhanced surrogates within a multi-fidelity context, with Han et al. (2013) proposing an additional hybrid bridge function to enhance multi-fidelity surrogates.

MSM has seen notable advancements in aerodynamic modelling and optimisation strategies. For instance, Mackman et al. (2013) compare various adaptive sampling methods for generating surrogate aerodynamic models applied in aerospace engineering. Kuya et al. (2011) extend this application to the multi-fidelity surrogate modelling of experimental and computational aerodynamic datasets, illustrating the technique's versatility. In terms of optimisation, the works of Forrester et al. (2006, 2007) and Eldred et al. (2004, 2006) play a significant role in integrating surrogate models with partially converged computational fluid dynamics simulations and developing formulations for surrogate-based optimisation with data fit and multifidelity models. Choi et al. (2005) applied a multi-fidelity framework to supersonic applications. These advancements enhance the reliability and efficiency of surrogate models and broaden their applicability in various engineering domains.

In computational modelling and data analysis, multi-fidelity surrogate modelling (MFSM) and data fusion have emerged as valuable interdisciplinary areas, addressing the challenges of data's increasing complexity and scale. The fundamental concepts, methodologies, and applications provide a foundational understanding and insight into its growing significance, as Fernández-Godino's comprehensive review from 2023 outlines. The review provides detailed insight into the evolution and current state of multifidelity (MM) and multifidelity surrogate models (MSM), highlighting their significance in computational sciences.

This study proposes a new fusion approach to construct multi-fidelity surrogate models by merely constructing gradient-only surrogates that use only gradients and regression. Although omitted in this study, the proposed approach is easily extended to weighted regression to accommodate the heteroscedastic nature of spatial variance resulting from multi-fidelity data. Wilke (2016) proposed gradient-only surrogates to construct smooth surrogates for discontinuous functions, and test problems were presented by Snyman and Wilke (2018). Gradient-only surrogates for multi-fidelity data are limited to line search approximations investigated by Chae and Wilke (2019) and Chae et al. (2023).

## 2. Surrogate Strategies

Radial basis function surrogates with randomly sampled centres are considered in this study. Three surrogate types are considered: those using only function (f) values, functions, and gradients (f-g), and only gradients (g). Surrogates constructed using only gradients are not unique, as any constant offset of the surrogates fits equally well. Therefore, all gradient-only surrogates are translated to 0 as the lowest value. All studies utilise Gaussian kernels, as formulated by Snyman and Wilke (2018). The shape parameter is uniformly sampled on a log scale between $10^{-4}$ and $10^5$ using 121 points. The shape parameter resulting in the lowest mean squared error on the training set is selected. Although prone to overfitting, we restrict the number of basis functions to be at least six times less than the number of observations, thereby mitigating the main risks of overfitting through regression, ensuring the ready reproduction of this study. For the interested reader, Snyman and Wilke (2018) and Correia and Wilke (2021) provide detailed discussions on improving the selection of the optimal shape parameter.

## 3. Experimental Setup

The foundational problem provides a simple yet effective illustration of the fundamental essence of the multi-fidelity problem. The study distils the core concepts and challenges associated with multi-fidelity modelling and optimisation by fitting a quadratic function. This example is a foundational reference point for understanding the broader implications of multi-fidelity approaches in more complex scenarios through the systematic variation of mini-batch sizes and their corresponding impact on function fitting and convergence. In addition to enhancing scientific reproducibility, it acts as a fundamental building block to help researchers and practitioners grasp the key principles and potential benefits of applying multi-fidelity techniques to various real-world problems.

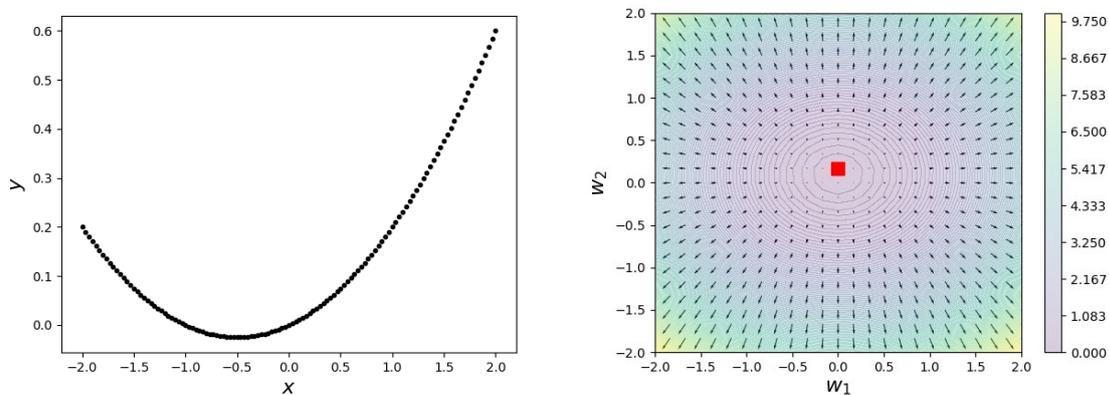

**Figure 2** The full-batch training dataset (left) and the sampled Mean Squared Error loss surface for varying $w_1$ and $w_2$ between -2 and 2 (right). The full batch loss surface is a baseline for the mini-batch sampled loss surfaces. The red square indicates the lowest sampled function value.

The foundational problem addressed in this study centres around fitting a one-dimensional quadratic function of the form $w_1 x^2 + w_2 x$. The full-batch training data is generated by sampling a quadratic function of the form, $0.1 x^2 + 0.1 x$, uniformly across the interval [-2, 2] with 121 data points, as depicted in Figure 2.

The main aim is to construct a surrogate function of the Mean Squared Error (MSE) loss surface when sampling $w_1$ and $w_2$ over a 25x25 full-factorial grid. Each variable $w_1$ and $w_2$ varies between [-2, 2]. The surrogate functions are constructed by subsampling the full-batch training data using stochastic mini-batches to evaluate the MSE loss function and gradients. This resembles differences in model fidelity as biases in function values and sensitivity changes in gradients are captured in the sampled data as expected in any multi-fidelity system. More samples resemble higher fidelity data, while fewer samples represent lower fidelity data. The mini-batches considered have a maximum batch size of 3 and 30, where the number of samples per batch is randomly sampled from 1 to the maximum batch size. It is important to note that only three sampled function values are required to recover the quadratic function exactly. This is readily achieved using an optimality criterion solution for the least-squares problem in closed form (see Snyman and Wilke (2018) for a detailed derivation). However, a sampling approach introduces changes in biases and sensitivities between [$w_1$, $w_2$] weights, introducing discontinuities (biases) and sensitivity changes (fidelity) between adjacent weights as expected from any multi-fidelity system. This renders the Mean Squared Error (MSE) loss surface discontinuous over $w_1$ and $w_2$, as shown in Figure 3.

Figure 3 depicts four figures as a 2x2 grid. Along the columns, this experimental computation uses mini-batches of varying maximum mini-batch sizes, specifically {3, 30} samples per batch. The maximum indicates the maximum sample per mini-batch uniformly sampled from 1 to the maximum mini-batch size. Each row represents a different set of samples, highlighting how the choice of mini-batch size impacts the behaviour of the MSE. In addition, a reference plot is provided at the bottom centre, representing the full batch loss surface, serving as a benchmark for comparison. This reference allows us to visually assess the performance of different mini-batch sizes relative to the full batch scenario. The red squares pinpoint the location with the lowest sampled function value.

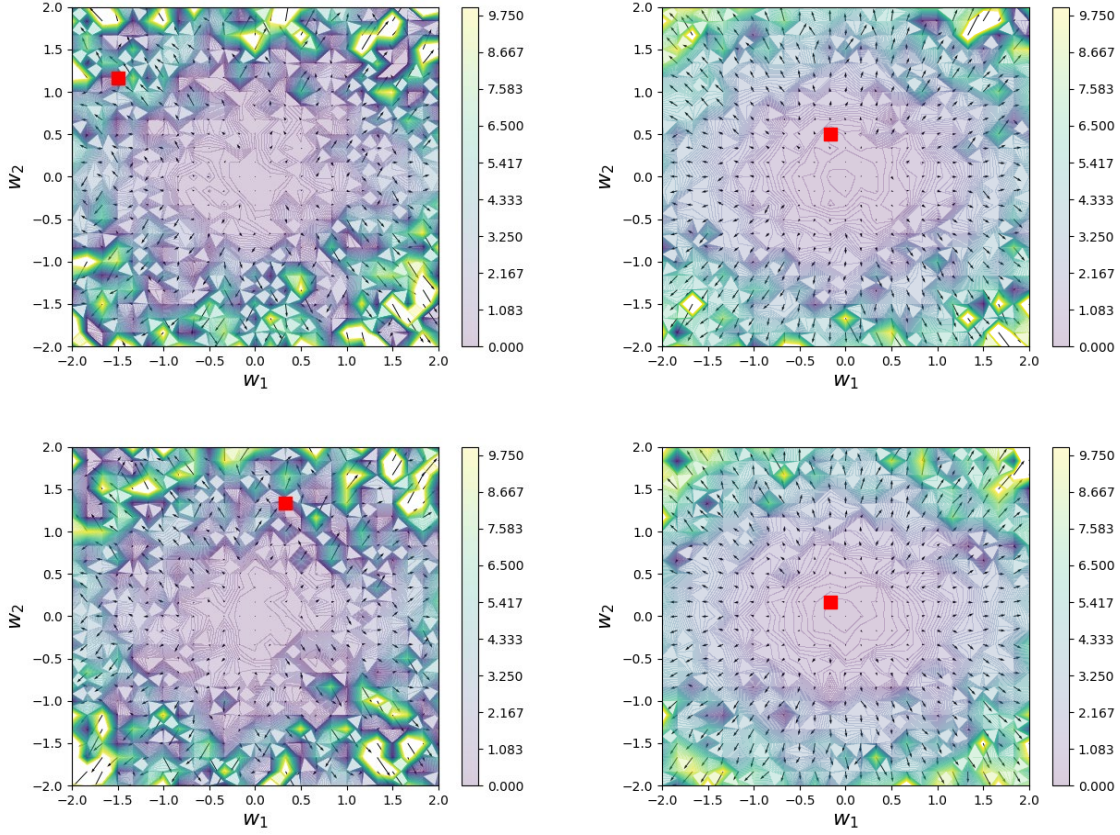

**Figure 3** Mean Squared Error (MSE) training data sampled over a 25x25 full factorial grid for mini-batches of maximum sizes {3, 30} (over the columns). Two sampled surfaces (over the rows) indicate typical function and gradient value changes between samplings. The full batch loss surface is included as a reference (bottom centre). The red square in each subplot indicates the lowest sampled function value.

## 4. Results

The study investigates the Mean Squared Error (MSE) loss surfaces in Radial Basis Function (RBF) surrogate construction, examining mini-batch sizes and RBF centre counts for simulating multi-fidelity systems. The MSE loss surfaces are analysed for mini-batches up to size three and RBF centres numbering 1 and 100, exploring methods 'f', 'f-g', and 'g' in surrogate construction. Each method distinctly impacts the MSE loss surface, as shown in Figures 4 and 5. Surrogates built using gradients alone ('g') are positive definite in all cases. In contrast, using function values alone ('f') or combined with gradients ('f-g') may lead to negative definite functions for mini-batch sizes of 3 and 30. Highlighting how additional data, while relevant, can degrade fit quality, introducing local minima into the convex loss function. Thus, careful data curation is vital to maintain surrogate quality. The evidence demonstrates the benefits of using only gradients as a multi-fidelity fusion strategy, rendering gradients a simple and scalable data curation strategy when readily available and accurately computed.

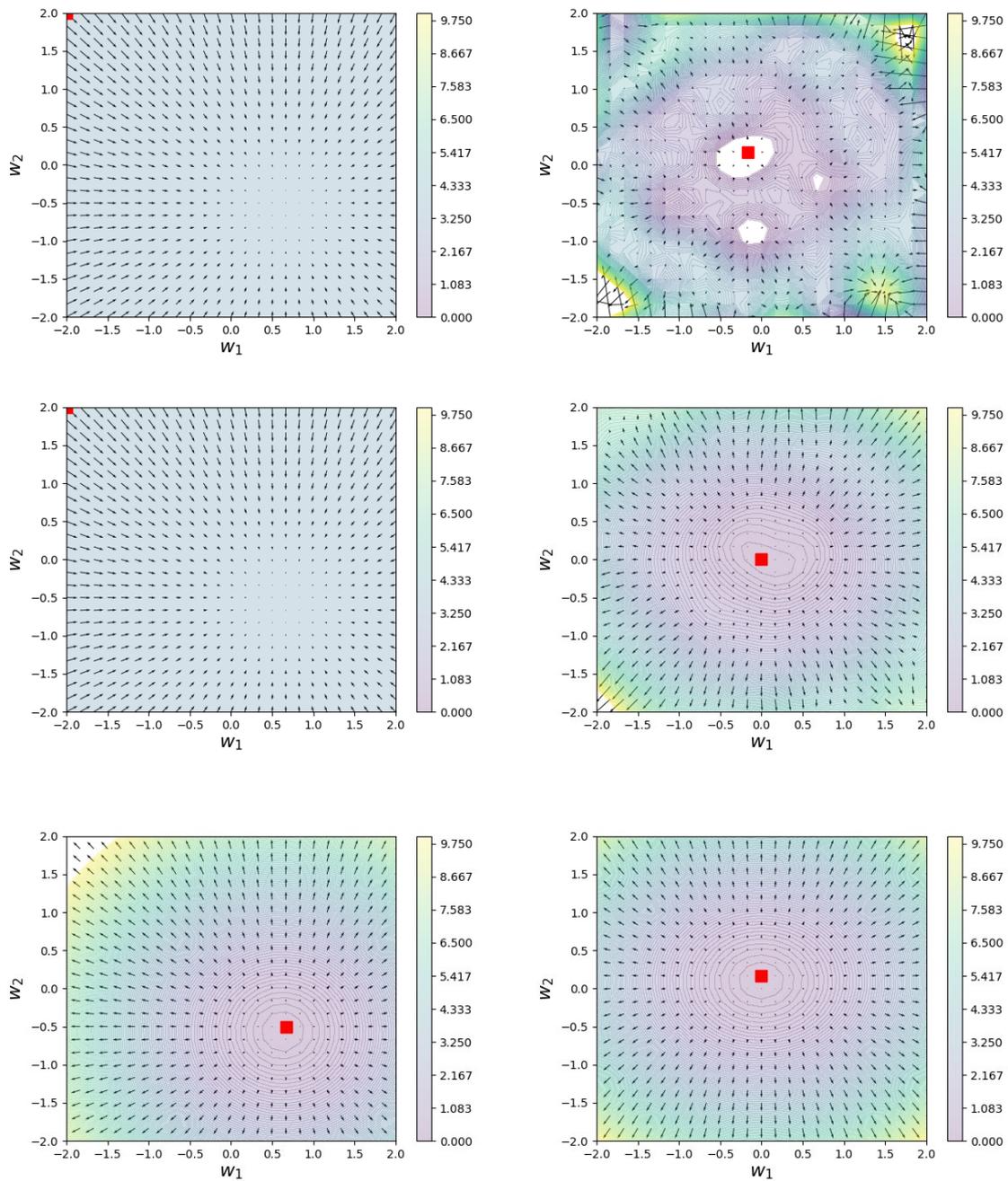

**Figure 4** Mean Squared Error (MSE) loss surfaces for mini-batches of maximum size {3} (randomly sampled), and {1, 100} RBF centres (over the columns) when constructing RBF surrogates using {f, f-g, g} (over the rows).

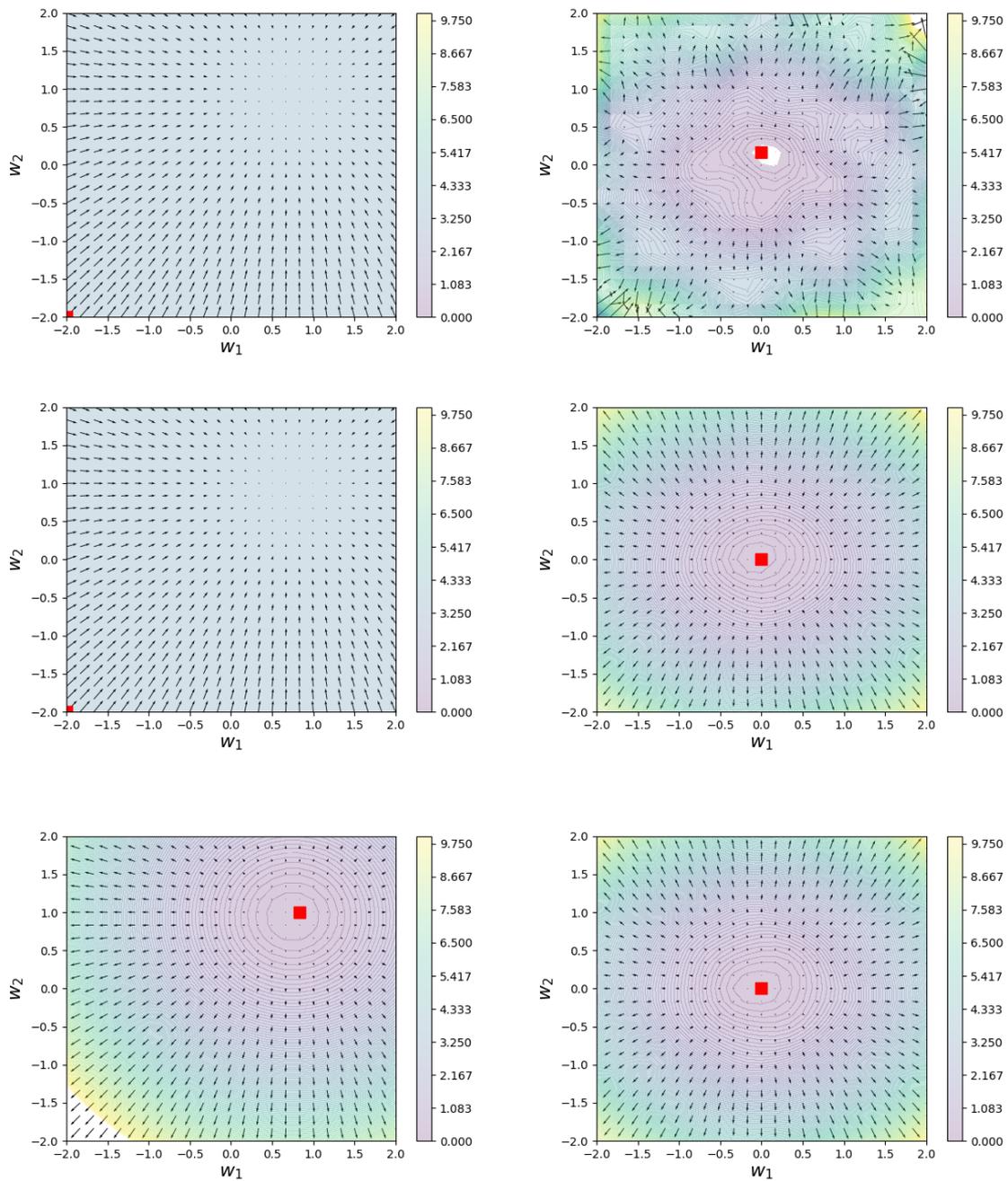

**Figure 5** Mean Squared Error (MSE) loss surfaces for mini-batches of maximum size {30} (randomly sampled), and {1, 100} RBF centres (over the columns) when constructing RBF surrogates with {f, f-g, g} (over the rows).

## 5. Conclusion

In conclusion, this study represents a significant leap forward in understanding multi-fidelity surrogate models, offering an innovative perspective of prioritising gradients and omitting function values. This approach's elegance is that multi-fidelity data fusion is achieved by constructing gradient-only surrogates. This key insight is demonstrated for a foundational problem involving the least squares fit of a basic quadratic function. The conventional notion that more data leads to better results in regression is challenged.

The perspective that gradient-only surrogates bring is that this research highlights the importance of discerning seemingly applicable data between high-quality and low-quality data. By adopting this refined approach, this study has demonstrated that the fusion of carefully curated, high-quality data can often outperform the inclusion of vaster quantities of uncurated yet seemingly applicable data. This shift in paradigm challenges the often conventional notion and underscores the critical role of data quality assessment and curation in developing and optimising surrogate models.

The main takeaway from this study is that achieving better results is not merely a matter of accumulating more data but rather of making informed decisions about the quality and relevance of the data. This profound insight opens new avenues for research and practical applications, emphasising the need to prioritise data relevance and quality over quantity. This will ultimately lead to more efficient and effective multi-fidelity modelling and optimisation strategies.

## References

Chae, Y., & Wilke, D. N. (2019). Empirical study towards understanding line search approximations for training neural networks. arXiv preprint arXiv:1909.06893.

Chae, Y., Wilke, D. N., & Kafka, D. (2023). Gradient-only surrogate to resolve learning rates for robust and consistent training of deep neural networks. Applied Intelligence, 53(11), 13741-13762.

Choi, S., Alonso, J., & Kroo, I. (2005). Multi-fidelity design optimization studies for supersonic jets using surrogate management frame method. In 23rd AIAA Applied Aerodynamics Conference, page 5077.

Correia, D., & Wilke, D. N. (2021). Purposeful cross-validation: A novel cross-validation strategy for improved surrogate optimizability. Engineering Optimization, 53(9), 1558-1573.

Eldred, M., Giunta, A., & Collis, S. (2004). Second-order corrections for surrogate-based optimization with model hierarchies. In 10th AIAA/ISSMO Multidisciplinary Analysis and Optimization Conference, page 4457.

Eldred, M., & Dunlavy, D. (2006). Formulations for surrogate-based optimization with data fit, multifidelity, and reduced-order models. In Proceedings of the 11th AIAA/ISSMO Multidisciplinary Analysis and Optimization Conference, number AIAA-2006-7117, Portsmouth, VA, volume 199.

Fernández-Godino, M. Giselle, Dubreuil, Sylvain, Bartoli, Nathalie, Gogu, Christian, Balachandar, Sivaramakrishnan, & Haftka, Raphael T (2019). Linear regression-based multifidelity surrogate for disturbance amplification in multiphase explosion. Structural and Multidisciplinary Optimization, 60, 2205–2220.

Fernández-Godino, M. G. (2023). Review of multi-fidelity models. Advances in Computational Science and Engineering, 1(4), 351-400.

Forrester, A. I., Bressloff, N. W., & Keane, A. J. (2006). Optimization using surrogate models and partially converged computational fluid dynamics simulations. Proceedings of the Royal Society of London A: Mathematical, Physical and Engineering Sciences, 462(2071), 2177–2204.

Forrester, A. I., Sóbester, A., & Keane, A. J. (2007). Multi-fidelity optimization via surrogate modelling. Proceedings of the Royal Society of London A: Mathematical, Physical and Engineering Sciences, 463(2088), 3251–3269.

Han, Z.-H., Görtz, S., & Zimmermann, R. (2013). Improving variable-fidelity surrogate modeling via gradient-enhanced kriging and a generalized hybrid bridge function. Aerospace Science and Technology, 25(1), 177–189.

Kuya, Y., Takeda, K., Zhang, X., & J. Forrester, A. I. (2011). Multifidelity surrogate modeling of experimental and computational aerodynamic data sets. AIAA journal, 49(2), 289–298.

Lee, S., Kim, T., Jun, S. O., & Yee, K. (2016). Efficiency enhancement of reduced order model using variable fidelity modeling. In 57th AIAA/ASCE/AHS/ASC Structures, Structural Dynamics, and Materials Conference, page 0465.

Mackman, T., Allen, C., Ghoreyshi, M., & Badcock, K. (2013). Comparison of adaptive sampling methods for generation of surrogate aerodynamic models. AIAA journal, 51(4), 797–808.

Robinson, T., Eldred, M., Willcox, K., & Haimes, R. (2008). Surrogate-based optimization using multifidelity models with variable parameterization and corrected space mapping. AIAA Journal, 46(11), 2814–2822.

Singh, G. and Grandhi, R. V. (2010). Mixed-variable optimization strategy employing multifidelity simulation and surrogate models. AIAA journal, 48(1):215–223.

Snyman, J. A., & Wilke, D. N. (2018). Practical Mathematical Optimization: Basic Optimization Theory and Gradient-Based Algorithms (Springer Optimization and Its Applications Series). Springer Cham.

Toal, D. J. (2015). Some considerations regarding the use of multi-fidelity Kriging in the construction of surrogate models. Structural and Multidisciplinary Optimization, 51(6), 1223–1245.

Wilke, D. N. How to get rid of discontinuities when constructing surrogates from piece-wise discontinuous functions. In J. Herskovits, editor, 5th International Conference on Engineering Optimization, Iguassu Falls, Brazil, 2016.

Yamazaki, W., Rumpfkeil, M., & Mavriplis, D. (2010). Design optimization utilizing gradient/Hessian enhanced surrogate model. In 28th AIAA Applied Aerodynamics Conference, page 4363.

Zimmermann, R., & Han, Z. (2010). Simplified cross-correlation estimation for multi-fidelity surrogate Cokriging models. Advances and Applications in Mathematical Sciences, 7(2), 181–202.

Zhang, Y., Kim, N.H., Park, C., & Haftka, R. T. (2018). Multifidelity surrogate based on single linear regression. AIAA Journal, 56(12), 4944–4952.